\documentclass[suppldata]{interact}

\usepackage{caption}
\usepackage[table]{xcolor}
\usepackage{booktabs}
\usepackage{multirow}
\usepackage{graphicx} 
\usepackage{amsfonts} 

\usepackage{epstopdf}
\usepackage{subfigure}

\usepackage{natbib}
\bibpunct[, ]{(}{)}{,}{a}{}{,}
\renewcommand\bibfont{\fontsize{10}{12}\selectfont}

\theoremstyle{plain}

\theoremstyle{definition}

\theoremstyle{remark}

\begin{document}


\title{{\itshape PlanBench-V}: 
A Spatial Planning Map Benchmark
for Vision-Language Models}

\author{
Minxin Chen\textsuperscript{*,1,2},
He Zhu\textsuperscript{*,1,2},
Junyou Su\textsuperscript{1,2},
Wen Wang\textsuperscript{1,2},\\
Yijie Deng\textsuperscript{1,2},
Wenjia Zhang\textsuperscript{\textdagger,1,3}\thanks{$^\ast$Equal contribution. \quad $^\dagger$Corresponding author. Email: wenjiazhang@tongji.edu.cn \vspace{6pt}}
\\[2ex]
\textsuperscript{1}Behavioral and Spatial AI Lab, Tongji University\\
\textsuperscript{2}Behavioral and Spatial AI Lab, Peking University \\
\textsuperscript{3}College of Architecture and Urban Planning, Tongji University
\\[2ex]
}

\maketitle

\begin{abstract}
Spatial planning maps play a central role in territorial governance by translating planning objectives, regulatory frameworks, and spatial strategies into visual representations. These maps serve not only as technical instruments for decision-making but also as key media for public communication and institutional coordination. However, interpreting spatial planning maps requires a combination of fine-grained visual perception, spatial reasoning, and policy-informed professional judgment, posing substantial challenges for both human learners and artificial intelligence systems.
With the rapid development of Vision-Language Models (VLMs), their potential application in urban planning analysis has attracted increasing attention. Nevertheless, existing multimodal benchmarks largely focus on general visual understanding and fail to capture the domain-specific cognitive processes embedded in planning practice. In this work, we introduce \textbf{PlanBench-V}, the first comprehensive benchmark designed to evaluate VLM performance in spatial planning map interpretation. First, we construct the Spatial Planning Map Database (SPMD), an expert-annotated dataset containing 223 planning maps and 1629 question--answer pairs curated by professional planners, covering diverse geographic regions and cartographic styles. Second, we propose a theory-informed evaluation framework that assesses model capabilities across four progressive dimensions: Perception, Reasoning, Association, and Implementation, reflecting the full cognitive pipeline of planning map interpretation.
Extensive experiments across two generations of VLMs reveal a clear trajectory of progress, yet also persistent challenges. While the best 2026 agentic reasoning model, Qwen3.6-Plus, substantially outperforms the best 2025 model, GPT-4o, by a margin of 27\%, all models continue to struggle with implementation-oriented tasks that require evaluative judgment, policy sensitivity, and constraint-aware decision-making. Our findings highlight fundamental limitations of existing VLMs in professional planning contexts and underscore the need for domain-adaptive multimodal reasoning frameworks. Code and data are available at \texttt{https://plangpt.github.io/}.
\end{abstract}


\section{Introduction}
Spatial Planning maps serve as essential tools in urban development, functioning as socioeconomic and thematic visualizations that document existing conditions, illustrate future scenarios, and guide policy implementation. 
Through specialized cartographic expressions, planners communicate vision, priorities, and constraints using symbols and annotations that convey land use allocations, infrastructure layouts, and functional zoning. These maps employ geographic space as a structured canvas, depicting planning elements with specific scales and orientations to ensure clarity in spatial arrangements. 
Unlike general-purpose maps, planning maps feature distinct representation elements and vary from large-scale master plans to detailed urban designs, each with unique stylistic features \citep{Lynch1984, Steinitz1995, Healey1997}. This study focuses specifically on \emph{statutory} spatial planning maps, as broader categories such as conceptual diagrams or urban design illustrations are not always tied to explicit, regulated planning intentions.
Enabling VLMs to effectively interpret these specialized planning maps would significantly enhance both professional practice and educational contexts in urban planning.

While VLMs have demonstrated impressive capabilities in general vision-language tasks, their effectiveness in highly specialized domains remains largely unexplored. Planning maps embody a high degree of complexity and professional specificity, integrating fine-grained visual elements (such as symbols, legends, and color codes), intricate spatial structures and layout relationships, and planning semantics closely tied to regulatory and policy frameworks. However, current VLMs often struggle to fully recognize essential cartographic elements like map legends, geographic entities, and planning boundaries. They also exhibit limited ability to parse complex spatial relations or infer the embedded planning logic and spatial strategies. Furthermore, VLMs generally lack a deep understanding of domain-specific terminologies and formal planning expressions, which impedes accurate interpretation and normative reasoning.

In addition, spatial planning has increasingly emphasized human-centered governance and public participation. The shift from technocratic planning toward participatory approaches highlights the growing importance of ensuring that planning information is accessible to diverse audiences, including residents, developers, policymakers, and researchers. Yet current VLMs provide insufficient support for non-expert users, thereby limiting public engagement and transparency. Their inability to effectively convey planning intentions across stakeholder groups also weakens interdisciplinary collaboration and hampers the development of intelligent, data-driven planning support systems. If VLMs are to function as mediators between technical planning documents and broader audiences, they must be capable of accurately interpreting planning maps while respecting institutional and regulatory contexts. The absence of systematic evaluation frameworks hinders progress toward this goal.

To address this gap, we introduce \textbf{PlanBench-V}, a benchmark specifically designed to assess the planning map understanding capabilities of VLMs. PlanBench-V consists of two main components: a high-quality dataset and a domain-informed evaluation framework.

The dataset comprises 223 spatial planning maps, derived from (1) official master plans in China; (2) {\it Chinese Certified Urban-Rural Planner Qualification Examination (CURPQE)}; and (3) international spatial planning maps, to ensure both accuracy and diversity in visual styles. Complementing this, we construct 1629 question-answer pairs manually annotated by professional urban planners, providing a rigorous basis for evaluating reasoning grounded in real-world planning practice.

\begin{figure}[htbp]
  \centering
  \includegraphics[width=0.8\linewidth]{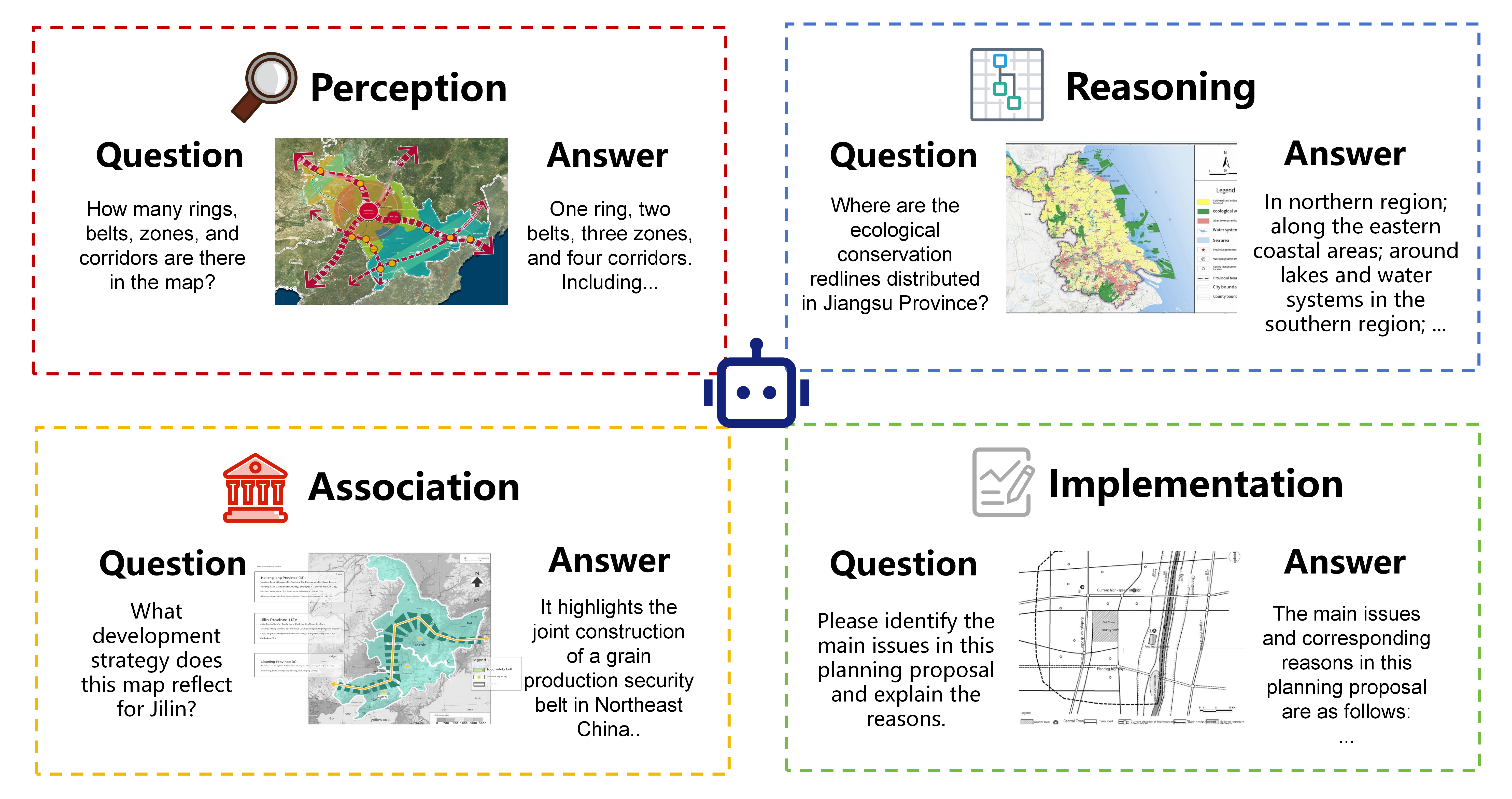}
  \centering
  \caption{Overview of the proposed PlanBench-V} 
  \label{fig:benchmark}
\end{figure}

We developed a comprehensive benchmark to systematically assess the capabilities of VLMs in understanding spatial planning maps. As illustrated in Figure~\ref{fig:benchmark}, this framework is structured across four key dimensions:

\begin{itemize}
    \item \textbf{Perception}: evaluates models' ability to identify visual components, including layout configurations, textual annotations, basic geographic features, and drawing elements.
    
    \item \textbf{Reasoning}: focuses on how well the model can derive structured insights and perform domain-specific reasoning from the complex visual and semantic information embedded in planning maps.
    
    \item \textbf{Association}: assesses the ability to collect and relate background policies, regulations, and planning indicators relevant to planning maps.
    
    \item \textbf{Implementation}: addresses the capacity for comparing, critiquing, and optimizing planning proposals.
\end{itemize}

In summary, our contributions are as follows:

\begin{enumerate}
    \item We introduce \textbf{PlanBench-V}, the first benchmark explicitly designed to evaluate vision-language models in the context of spatial planning map interpretation, grounded in real-world planning documents and expert knowledge.

    \item We propose a four-dimensional assessment framework---Perception, Reasoning, Association, and Implementation---that reflects the progressive cognitive structure of planning practice, bridging foundational map literacy and applied professional judgment.

    \item Through extensive evaluation across two generations of VLMs, we reveal substantial performance gains driven by agentic reasoning, yet also persistent gaps in implementation-oriented tasks that require policy sensitivity and evaluative judgment, offering guidance for future domain-adaptive multimodal model development.
\end{enumerate}

\section{Related Work}
\subsection{Spatial Planning Maps}
Spatial Planning maps represent a specialized form of cartographic visualization that has evolved alongside urban planning practice. 
They are not merely technical drawings of space; rather, they are symbolic representations that communicate planning intentions, values, and ideologies within specific political and institutional contexts \citep{harley1989deconstructing, harley2002new, albrechts2003strategic, duhr2007visual}. These maps are integral to spatial planning processes, functioning as tools of communication, negotiation, and persuasion. 

Spatial planning maps can be broadly categorized according to their scale, purpose, and legal status. At the national and regional level, strategic spatial maps provide high-level visions for spatial development, often emphasizing economic corridors, ecological networks, and urban hierarchies \citep{faludi2000esdp, albrechts2003strategic}. At the municipal level, zoning maps delineate specific regulations regarding land use, density, and building form \citep{burgess2009shift}. Between these extremes, a range of non-statutory or illustrative maps exists, including conceptual diagrams, master plans, and urban design frameworks that aim to foster public engagement and design communication \citep{duhr2009visualising, li2025communicative}.

The cartographic design of these maps presents distinctive challenges in encoding abstract policies into visual form. This process involves careful decisions about symbol systems, visual hierarchy, and spatial composition. The semiotic dimension, encompassing visual grammar, symbolization, and layout, serves as the foundation for making complex planning intentions interpretable to various audiences \citep{bertin1983semiology, duhr2007visual}. Complementing this, analytical typologies classify maps into descriptive, evaluative, and prescriptive forms, each serving specific functions: documenting existing conditions, assessing potentials, or guiding future spatial organization \citep{moroni2017graphic}.

The interpretation of spatial planning maps relies on a combination of perceptual and cognitive mechanisms that are grounded in both innate visual processing and acquired knowledge structures. At the perceptual level, users extract immediate visual cues, such as shape, size, proximity, and contrast, that are essential for recognizing zoning patterns, infrastructure alignments, and land-use hierarchies \citep{palmer1999vision, rautenbach2017Developmentevaluation}. Gestalt principles like proximity, continuity, and figure-ground differentiation play a central role in how viewers form holistic spatial understandings from complex visual data \citep{wertheimer1938gestalt, koffka1935principles}. Beyond perception, cognitive processes such as schema activation and pattern recognition enable users to interpret map symbols, like colors, lines, and textures, based on prior knowledge and contextual expectations \citep{mayer2005cambridge, arnheim1969visual}. By leveraging both visual and cognitive principles, planners can ensure that spatial representations are not only technically accurate but also intuitively accessible across diverse stakeholder groups \citep{kress2006reading, alexander1977pattern}.

Despite the extensive understanding of spatial planning maps, gaps remain in the cognitive processes involved in map interpretation, particularly regarding how different stakeholders decode and utilize these maps. 



\subsection{VLMs and Benchmarks in Specialized Domains}

VLMs have demonstrated remarkable progress in general multimodal tasks, including image understanding \citep{gpt4o,gemini}, visual reasoning \citep{zhu2025internvl3,guo2025seed1}, and multimodal dialogue \citep{liu2023llava, wang2024qwen2vl}. Recent research has successfully extended these models to specialized domains such as medical imaging \citep{li2023llava, lai2025med, pan2025medvlm}, geographical information systems \citep{zhang2024bb, zhang2024mapgpt}, and mathematical reasoning \citep{chen2025bring, shen2025vlm}.
The development of comprehensive benchmarks has been crucial for advancing multimodal AI capabilities. Established datasets like VQA \citep{agrawal_vqa_2016}, COCO-Captions \citep{lin_microsoft_2015}, and Visual Genome \citep{krishna_visual_2017} have driven progress in general vision-language understanding. More recent specialized benchmarks have emerged for specific domains, including scientific diagrams \citep{li2023scigraphqalargescalesyntheticmultiturn,roberts2024scifibenchbenchmarkinglargemultimodal}, document understanding \citep{mathew2021documentvisualquestionanswering,wang_needle_2024}, aesthetics \citep{huang2024aesbench, zhou2024uniaa, lin2024designprobe}, autonomous driving \citep{qian2024nuscenes, sima2024drivelm}, and other fields.

In the geographical and cartographic domain, MapQA \citep{chang2022mapqadatasetquestionanswering} and the Charting New Territories dataset \citep{roberts2023charting} provided an initial evaluation for map-based question answering, such as localization and identification. However, these resources primarily address basic geographic interpretation rather than the complex domain-specific requirements of planning map analysis.

We identify urban planning as a critical domain that could significantly benefit from specialized VLMs to \textit{interpret complex planning maps}---a task where even leading commercial models exhibit substantial limitations in recognizing specialized elements and applying the cartographic interpretation skills essential for planning practices. Recent efforts toward domain-tailored systems, such as PlanGPT \citep{zhu-etal-2025-plangpt}, which integrates customized retrieval, domain-specific knowledge activation, and tool orchestration to support urban planning workflows, and its multimodal extension PlanGPT-VL \citep{zhu2025plangptvlenhancingurbanplanning}, illustrate the value of planning-aware model design. However, these systems mainly target text generation or general visual understanding rather than the structured interpretation of planning maps, leaving the evaluation of map-grounded perception, reasoning, association, and implementation largely open---a gap that PlanBench-V is designed to fill.

\section{Constructing Database}
Understanding planning maps is an inherently abstract and complex task due to their highly specialized visual and semantic nature \citep{zhu2025plangptvlenhancingurbanplanning}. To ensure the reliability of our benchmark, the construction of a high-quality, professionally annotated planning map dataset is essential. This SPMD was meticulously curated by experts with domain-specific knowledge in urban planning and geography.

Our data synthesis pipeline consists of three core stages: parsing, question generation, and answer generation.

\begin{figure}
\centering
\includegraphics[width=0.8\linewidth]{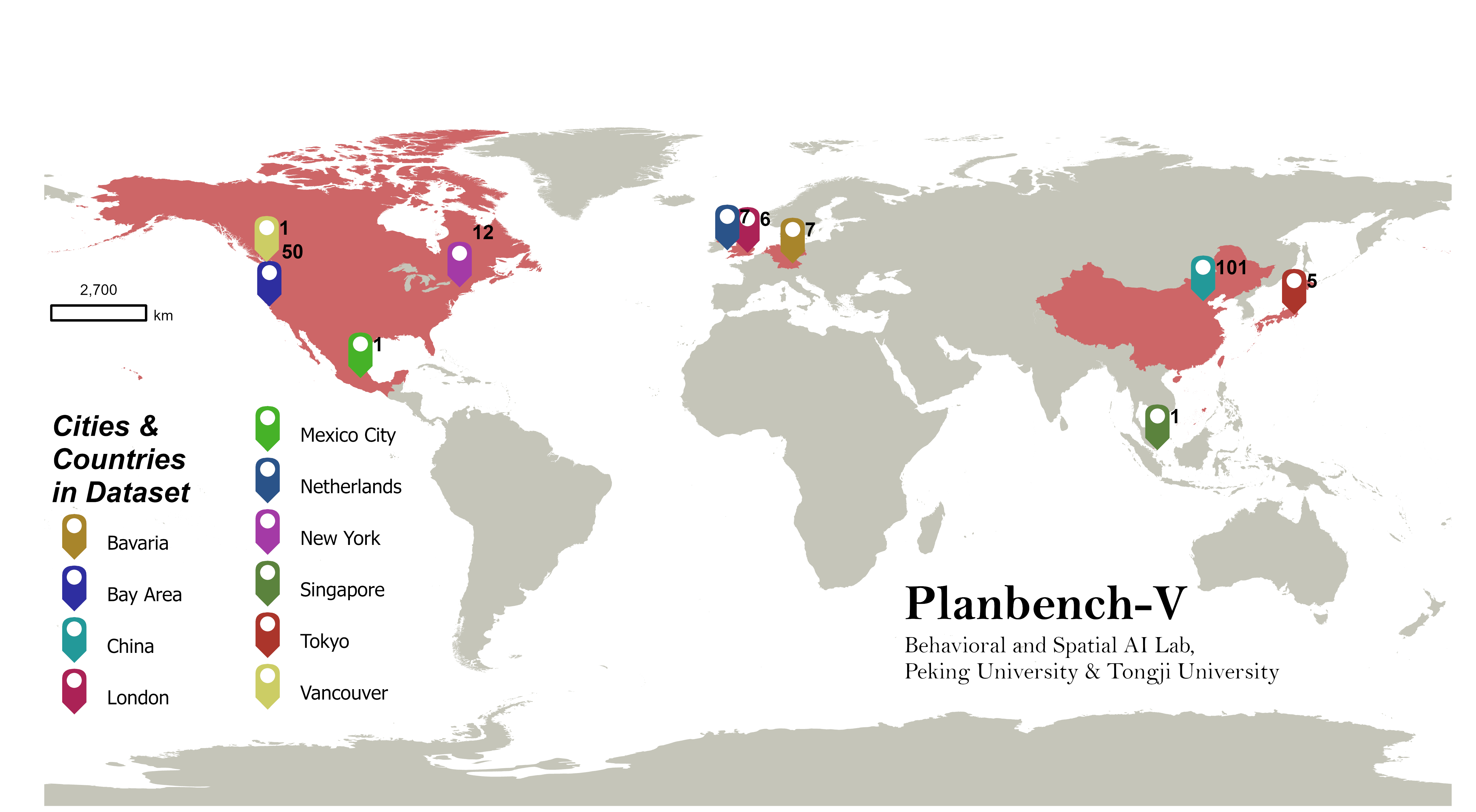}
\caption{Geographical Distribution of Planning Cases in SPMD}
\label{fig:distribution}
\end{figure}

First, in the parsing stage, we employed custom scripts to extract planning maps from official planning documents sourced from municipal governments, planning agencies, and academic institutions. Chinese planning documents were primarily collected from the official websites of provincial and municipal natural resource bureaus and from the \textit{CURPQE}. The spatial planning maps from other countries encompass a diverse range of cases as shown in Figure~\ref{fig:distribution}. 
The dataset includes three general and two thematic planning maps from Japan’s \textit{Tokyo 2040 Plan}; four maps from the UK’s \textit{London 2021 Master Plan} and two from the \textit{London Greenway}; and seven maps from the US \textit{New York 2050 Master Plan}, five thematic plans, and 50 additional maps from US cities, including San Francisco, Oakland, and Chicago. The Netherlands contributes seven maps from the \textit{The Mosaic} Master Plan, Canada provides one site plan from the \textit{Granville Island Redevelopment Project} in Vancouver, and Germany offers seven maps focused on village revitalization in Bavaria. Additionally, one planning map each is included from Singapore and Mexico.
Given that many documents contained extraneous visual elements such as icons, background images, and real-world photographs, we conducted manual screening to isolate clean planning maps. Where necessary, we enhanced low-resolution maps through sharpening or source tracing to ensure image clarity and legibility of embedded text.

Building on these parsed maps, a panel of experts with professional backgrounds in urban planning and geography, including 8 graduate-level students trained in advanced urban design and 1 practitioner with over four years of experience at a planning institute, formulated a set of questions that targeted key dimensions of visual–spatial reasoning in planning scenarios. These questions were explicitly designed to evaluate model capabilities across four core dimensions of planning map understanding: perception, reasoning, association, and implementation.

To minimize hallucination and ensure factual correctness, reference answers were generated through a combination of automated processes and expert validation. Specifically, answers initially generated by Qwen2.5-VL-72B-Instruct were reviewed, corrected, and rewritten by professional planners to produce authoritative references. To improve the ecological validity and generalizability of the benchmark, the dataset incorporates diverse planning styles and visual conventions, reflecting a variety of regional practices and map-making traditions. These efforts collectively enable more robust and reliable evaluation of VLMs' capabilities in real-world, expert-level planning contexts.

In total, we collected 223 images and 1629 QA pairs, as summarized in Table~\ref{tab:spmd_stats}. Specifically, these VQA pairs assess capabilities in perception (25.7\%), reasoning (54.5\%), association (18.0\%), and implementation (1.8\%). The visual materials cover diverse planning contexts, drawing on both Chinese sources (China Master Plans and \textit{CURPQE}) and international cases from Asian, European, and North-American cities.

\begin{table}[htbp]
\centering
\scriptsize
\caption{Statistical Information of SPMD}
\label{tab:spmd_stats}
\begin{tabular}{llrrrr}
\toprule
\textbf{Capability Level} & \textbf{Task} & \textbf{Total} & \textbf{CMP\textsuperscript{a}} & \textbf{CURPQE} & \textbf{Others\textsuperscript{b}} \\
\midrule
\multirow{2}{*}{Perception} 
  & Element Recognition & 317 & 165 & 52 & 100 \\
  & Caption             & 101 & 69  & 32 & 0   \\
\midrule
\multirow{3}{*}{Reasoning} 
  & Classification              & 294 & 147 & 47 & 100 \\
  & Spatial Relationship Reasoning & 300 & 149 & 52 & 99  \\
  & Domain-Specific Reasoning   & 294 & 141 & 53 & 100 \\
\midrule
Association 
  & Policy Association          & 294 & 143 & 50 & 101 \\
\midrule
\multirow{2}{*}{Implementation} 
  & Scheme Evaluation           & 20  & 0   & 20 & 0   \\
  & Decision-making             & 9   & 0   & 9  & 0   \\
\bottomrule
\end{tabular}
\centering
\tabnote{
\begin{minipage}{1\linewidth}
\textsuperscript{a}CMP stands for China Master Plans;
\textsuperscript{b}Spatial planning maps from Other countries, including Japan, United Kingdom, United States, Netherlands, Canada, Germany, Singapore, and Mexico.
\end{minipage}
}
\end{table}

\section{Developing the Metrics}
Our investigation begins with a comparative analysis of established benchmarks across professional image domains, including Engineering Documentation \citep{doris2024DesignQAMultimodal}, Graphic Design \citep{lin2024designprobe}, and Image Aesthetics Perception \citep{huang2024aesbench}. This review reveals the distinct nature of spatial planning maps, which occupy a unique position at the intersection of functional abstraction, symbolic representation, and domain-specific conventions. 

Unlike mechanical design drawings, which are strictly technical and follow standardized engineering protocols, or artistic images that prioritize aesthetic expression and subjective interpretation, planning maps balance functionality and minimal aesthetics. They employ simplified, codified, and standardized visual styles to depict land use, spatial layouts, transportation systems, and public facility allocations. Furthermore, unlike photographs that can be intuitively understood based on visual familiarity, planning maps require both perceptual recognition and abstract reasoning rooted in planning theory and regulatory frameworks, making their interpretation a highly specialized cognitive task.

To deconstruct this specialized task, we examined several foundational frameworks. Map reading  skills are outlined in Map Literacy frameworks \citep{rautenbach2017Developmentevaluation}, which identifies core competencies including: recognizing symbology, orientation and locating, measuring and estimating, calculating and explaining, and extracting knowledge. These skills form the mechanical basis for decoding the map's surface information. Transcending this, the ACRL Framework for Visual Literacy in Higher Education (2022) provides a higher-order cognitive lens, emphasizing critical interpretation through its four frames: perceiving visuals as information, practicing discernment and criticality, participating in the visual information landscape, and pursuing social justice through visual practice.

However, the ultimate measure of competence in urban planning lies in the application of this knowledge to solve real-world problems. The \textit{CURPQE} in China exemplifies this, particularly through its subjective questions on 'Urban Planning Practice'. These questions assess the core professional capabilities of a contemporary planner: the ability to solve complex problems and to apply established rules, technical standards, and legal principles to practical scenarios. This represents the synthesis of theoretical knowledge and interpretive skill into actionable professional judgment.


Based on this analysis, it is evident that a holistic assessment of planning map comprehension requires a framework that bridges foundational literacy with applied professional competence. We therefore propose a comprehensive evaluation framework constructed upon the urban planning knowledge system. The framework's logic follows a progressive path from theoretical foundations, through perceptual and interpretive mechanisms, to applied problem-solving capabilities. It specifically comprises the following four core dimensions:
\subsection{Perception}
The questions of the perception subset consist of 2 categories as follows:

\textbf{Element Recognition} evaluates models' ability to identify layout configurations, textual annotations, basic geographic features, and drawing elements in planning maps. It facilitates the establishment of semantic alignment between image content and natural language. This dimension includes a total of 317 VQA items, of which 165 are from the China Master Plan, 52 are from the \textit{CURPQE} and 100 from other countries.

\renewcommand{\arraystretch}{1.5} 
\begin{table}
\tbl{Example Questions on elements of Territorial Spatial Master Plan Maps\\
\textit{Refer to Specification for the Mapping of Municipal Territorial Spatial Master Plans in China}}
{\begin{tabular}{p{3cm}p{5cm}p{6cm}} \toprule
\textbf{Element} & \textbf{Content} & \textbf{Example Question(s)} \\ \midrule
Map Layout & Map Frame, North Arrow, Wind Rose, Scale Bar, Legend(Symbols) & Where is the north arrow located? What is the scale? \\
Textual Elements & Map Title, Credits and Date of Mapping, Explanatory Notes, Annotations, Labels, , Legend(text) & What does the legend include? Which government seats are annotated on the map? \\
Planning Elements & Base Map Elements: Administrative Boundary Features, Natural Geographic Features, Transportation Features, Land Use and Zoning Features; Mandatory Thematic Content Optional Thematic Content & What province is shown in the map? How many highways are there on the map? Which lakes and rivers are on the map? How many districts are in this province?What areas are included in the "three cores" shown in the map? \\ \bottomrule
\end{tabular}}
\label{tab:territorial-map-elements}
\end{table}

\renewcommand{\arraystretch}{1.3}  

\begin{table}
\tbl{Example Questions on spatial relationships of Territorial Spatial Master Plan Maps}
{\begin{tabular}{p{4.5cm}p{5cm}p{5cm}} \toprule
\textbf{Relationship} & \textbf{Meaning} & \textbf{Example Question} \\ \midrule
Topological Spatial Relationship & Relationships between entities such as adjacency, connectivity, inclusion, and intersection. & Are Nanjing and Yangzhou adjacent? What is the topological relationship between the Riverside Agricultural Zone and the Lixiahe Agricultural Zone? \\
Ordinal Spatial Relationship & Relationships in the distribution of spatial entities, described using directional terms such as up/down, left/right, front/back, and east/west/south/north. & In which parts of Jiangsu Province are the Ecological Protection Redlines mainly distributed? \\
Metric Spatial Relationship & Relationships concerning the distance (near or far) between spatial entities. & Is the Nanjing Metropolitan Area closer to the Su-Xi-Chang (Suzhou-Wuxi-Changzhou) Metropolitan Area or to the Huaihai Economic Zone? \\ \bottomrule
\end{tabular}}
\label{tab:spatial}
\end{table}

\renewcommand{\arraystretch}{1.3} 
\begin{table}
\tbl{Example Questions on domain-specific reasoning of Territorial Spatial Master Plan Maps}
{\begin{tabular}{p{4.5cm}p{5cm}p{5cm}} \toprule
\textbf{Aspect} & \textbf{Meaning} & \textbf{Example Question} \\ \midrule
Spatial Layout & Types of centralized urban layouts include grid, radial-ring, concentric, and fan-shaped forms. Dispersed layouts include cluster, constellation, star-shaped, linear, and circular types. & What are the characteristics of the province's territorial spatial layout? Why has such a layout formed? \\
Functional Organization & By rationally arranging urban functional zones, the spatial structure can be optimized to improve operational efficiency and residents' quality of life. & How is the province's territorial spatial functional organization implemented? How does it support the province’s socio-economic development? \\
Transportation System & Efficient transportation infrastructure and management ensure smooth movement of people and goods while reducing congestion and environmental pollution. & How does the province’s transportation system support its socio-economic development? What are the key features of its transportation planning and construction? \\
Environmental Ecology & Protecting and utilizing the natural environment and ecosystems enables sustainable urban development and enhances residents’ quality of life. & How does the province’s ecological and environmental planning ensure sustainable development? What are its priorities and challenges? \\ \bottomrule
\end{tabular}}
\label{tab:domain_specific}
\end{table}

\textbf{Caption}, in this study, refers to extracting as many details from the image as possible. The model generates descriptions based solely on the image itself, rather than identifying elements in response to a specific question. The quality of the caption reflects whether the model has truly "seen" the image clearly. This dimension includes a total of 101 VQA items, of which 69 are from the China Master Plan and 32 are from the \textit{CURPQE}.

An example Question would be: \textit{"Please describe this planning map in detail."}

\subsection{Reasoning}
This dimension encompasses classification, spatial relationship reasoning, and domain-specific reasoning.

\textbf{Classification}  focuses on the ability to recognize different types of planning maps. Due to the fact that this benchmark focuses on statutory spatial planning, other forms such as conceptual plans, urban design, district-scale strategic planning, and thematic studies are currently not included. Based on China’s “five-level, three-category” planning system, the maps are categorized into master plans, detailed plans (including both regulatory and site plans), and specialized planning documents. For maps without explicitly labeled types, such as those appearing in exams, experts annotated their types based on the scale of the map (e.g., national, provincial, municipal, or township level) and the visualized planning content. Maps presenting broad spatial structures without detailed regulatory indicators are generally considered master plans. Those containing control indicators such as building density, height limits, floor area ratio, green space ratio, and redline boundaries are identified as regulatory detailed plans. In contrast, maps that display site layouts or architectural schemes are classified as site plans, while maps focusing on specific domains or policy areas are categorized as specialized plans. This dimension includes a total of 294 VQA items, of which 147 are from the China Master Plan, 47 are from the \textit{CURPQE} and 100 from other countries.

\textbf{Spatial Relationship Reasoning}, as shown in Table~\ref{tab:spatial}, assesses the understanding of spatial relationships between geographic elements in planning maps, including: topological spatial relations, sequential spatial relations, and metric spatial relations. This dimension includes a total of 300 VQA items, of which 149 are from the China Master Plan, 52 are from the \textit{CURPQE} and 99 from other countries.

\textbf{Domain-specific Reasoning}, as shown in Table~\ref{tab:domain_specific}, encompasses four key aspects: spatial layout, functional organization, transportation system, and environmental ecology, each reflecting a critical dimension of professional interpretation in territorial spatial planning. This dimension includes a total of 294 VQA items, of which 141 are from the China Master Plan, 53 are from the \textit{CURPQE} and 100 from other countries.

\subsection{Association}
This dimension assesses the ability to collect and relate background policies and contextual documents relevant to planning maps. At a fine scale, it examines policy, regulations, and planning indicators.
In the third part of \textbf{PlanBench-V}, Association refers to the model's ability to retrieve, relate, and reason over background policies and contextual documents that are relevant to the content of planning maps. Unlike tasks that focus on visual identification or structural interpretation, association requires models to connect spatial elements with appropriate policy frameworks, including administrative hierarchies, regulatory classifications, and land-use guidelines. This task evaluates whether the model can "understand" not only what is shown on the map, but also the broader institutional and legal context in which the design operates.

At a finer level, this dimension includes tasks involving national and local planning policies, special-purpose plans, and technical reports. The model is expected to recognize references to zoning regulations, ecological protections, urban renewal standards, and policy constraints implicitly indicated by visual symbols or layout patterns. A strong performance reflects the model’s capacity to bridge visual and textual modalities and to align spatial content with normative planning knowledge.

This dimension includes a total of 294 VQA items, of which 143 are from the China Master Plan, 50 are from the \textit{CURPQE} and 101 from other countries.

An example question would be:
"Based on the content of this planning map, which relevant policies or guidelines should be considered when evaluating its implementation?"

\subsection{Implementation}
This dimension addresses the capacity for comparing, critiquing, and optimizing planning proposals. Unlike tasks focused on recognition or factual recall, Implementation requires models to make evaluative judgments under real-world constraints, such as balancing ecological protection, land-use efficiency, and development goals. This dimension reflects whether a model can move beyond descriptive outputs and generate insights that are strategic, selective, and professionally grounded.

Given the integrative nature of urban planning, models are expected not only to identify problems in design proposals but also to reason through trade-offs and suggest actionable revisions. Strong performance in this dimension indicates a grasp of planning logic, an ability to critique spatial solutions, and the use of terminology aligned with professional practice.

The questions of the Implementation subset consist of 2 categories as follows:

\textbf{Scheme Evaluation} measures the model's ability to evaluate the strengths and weaknesses of a planning proposal based on the visual input and spatial context. Responses are expected to reflect planning-specific criteria such as land use distribution, spatial continuity, accessibility, and compatibility. This subset includes 20 VQA items, all derived from the \textit{CURPQE}, and presents both schematic diagrams and explanatory texts. Example question: “What are the strengths and weaknesses of this planning scheme?”

\textbf{Decision making} tests whether the model can make value-driven choices in a constrained design scenario. The model must compare multiple alternatives or identify optimal directions based on planning principles and contextual information. It includes 9 VQA items, all from \textit{CURPQE}, often accompanied by prompts requiring the selection or justification of preferred options.

Example question:
“Which planning direction is more suitable for the site, and why?”

\section{Experiments}

\subsection{Experimental Setup}
\label{sec:exp_setup}

We conducted extensive experiments on 17 VLMs to evaluate their capabilities in interpreting planning maps. The first round includes GPT-4o and GPT-4o-mini together with nine state-of-the-art variants from the popular Intern \citep{chen2023internvl} and Qwen-VL \citep{bai2023qwen} model families, while the second round adds six agentic reasoning models. To ensure completeness and fairness, all VLMs were evaluated using their original released weights without any dataset-specific fine-tuning.

Qwen and InternVL, especially 2.5 and 3-series models, were selected due to their strong performance as open-source models in existing multimodal benchmarks. MMMU \citep{yue2023mmmu} consists of 11.5k multimodal questions derived from university-level course content, and MMBench v1.1 covers a wide range of disciplines from fundamental sciences to engineering applications. These benchmarks motivated the inclusion of Qwen2.5-VL-72B, InternVL2.5-78B-MPO, and InternVL3 variants as representative open-source baselines.

To systematically analyze the impact of model scaling laws, our selection spans a full range from lightweight to flagship models. Within the first-round Qwen series, we included models with 2B, 3B, 7B, and 72B parameters. In the InternVL3 series, we selected models with 8B, 9B, and 14B parameters. This broad parameter spectrum provides a solid empirical basis for investigating the trade-off between performance and resource consumption, as well as how different capability dimensions (e.g., perception, reasoning) evolve with model size. Crucially, to assess performance under practical deployment scenarios, we also included models that employ AWQ quantization, namely Qwen2-VL-72B-Instruct-AWQ and Qwen2.5-VL-72B-Instruct-AWQ. These quantized entries provide a practical view of compressed flagship-scale models alongside smaller non-quantized models.

Our evaluation methodology addresses the inherent complexity of planning problems by accommodating multiple valid approaches to the same challenge. Rather than enforcing a single correct answer, our framework evaluates responses based on adherence to planning principles, logical consistency, evidence-based reasoning, and consideration of diverse stakeholder perspectives. For bilingual evaluation, we developed specialized protocols that account for cross-cultural variations in planning terminology, regulatory frameworks, and professional practices.

\subsection{Exploration of Prompt Refining}
\label{sec:prompt_refining}

Given that the benchmark tasks primarily consist of subjective, judgment-intensive questions, we adopted the LLM-as-a-judge evaluation framework enhanced with structured scoring rubrics and automated prompt optimization. To establish a human performance baseline and validate the reliability of automated evaluation, we recruited three graduate students specializing in urban planning to answer 40 questions sampled across all eight task types (five questions per type). Their responses were independently scored by four LLM judges—GPT-4o-mini, GPT-4o, Claude Opus 4.7, and GPT-5.4—using identical rubrics.

\subsubsection{Human Performance Baseline and Judge Calibration}

A clear strictness gradient emerges: GPT-4o-mini is the most lenient judge (1.377/2, 68.9\%), followed by GPT-4o (1.342/2, 67.1\%), Claude Opus 4.7 (1.289/2, 64.5\%), and GPT-5.4 (1.255/2, 62.8\%). The per-item Spearman rank correlations among all judge pairs exceed 0.75 ($p < 0.001$ for all pairs), indicating strong agreement on the relative quality ordering of human responses. The four judges naturally partition into two clusters: GPT-4o-mini and GPT-4o form a more lenient group ($\rho = 0.894$), while Claude Opus 4.7 and GPT-5.4 form a stricter group ($\rho = 0.921$). GPT-4o occupies a favorable intermediate position, exhibiting high cross-cluster correlations (0.856 and 0.790 with Claude Opus 4.7 and GPT-5.4, respectively).

At the per-type level, correlations are moderately lower, with the weakest agreement observed between GPT-4o-mini and GPT-5.4 ($\rho = 0.548$, not significant), driven primarily by task categories such as Decision-making and Classification, where the two models apply substantially different evaluative standards. These results confirm that, while LLM judges are well-calibrated for within-model ranking, absolute score interpretation should account for judge-specific strictness.

Critically, the adoption of structured reference answers with annotated critical points—rather than open-ended rubrics—substantially narrows the gap between lenient and strict judges. When scoring prompts are anchored to predefined evaluation criteria, even the most cost-efficient model, GPT-4o-mini, achieves per-item correlations above 0.89 with GPT-4o and above 0.75 with the strictest judge, GPT-5.4. This establishes a practical cost--accuracy tradeoff: deploying GPT-4o-mini with well-designed critical-point rubrics yields evaluation fidelity comparable to that of substantially larger judges, at a fraction of the inference cost.

\subsubsection{Prompt Optimization and Scoring Protocol}

Our evaluation system was trained on expert-annotated planning assessments and incorporates structured scoring rubrics to enable consistent and reliable evaluation across diverse planning subdomains. To mitigate potential biases in subjective scoring, we developed a prompt optimization mechanism that iteratively aligns automated judgments with ratings provided by human experts. Experimental results demonstrate that the proposed method achieves an average pairwise error below 0.1 between human evaluators and the automated system, indicating high reliability even for complex, subjective planning tasks.

We conducted evaluations using GPT-4o-mini (temperature=0) as the primary evaluation model. To address variability introduced by API connection instability, we performed multiple evaluation runs per test case and averaged the resulting scores. For text-based tasks, we employed exact match or semantic similarity metrics for objective questions, while relying on the LLM-as-judge methodology with structured rubrics for subjective questions. Visual tasks were analogously partitioned into caption-based questions, which target descriptive analysis, and non-caption questions, which target spatial reasoning and planning interpretation.

We began by designing and testing multiple versions of evaluation prompts, each varying in phrasing, structure, and the granularity of evaluation dimensions. These prompt variants were applied across a diverse set of planning QA tasks, and their outputs were compared against expert-annotated ground truths. Empirically, we observed that stronger-performing VLMs tend to be more robust to prompt variation, yielding relatively stable scores across different evaluation formats.

However, we also identified several challenges during prompt optimization. We experimented with a range of scoring prompts spanning from minimal formulations to highly specific ones that emphasize geographic location, functional zoning, infrastructure layout, and element-to-legend alignment. When an excessive number of evaluation dimensions were introduced—for instance, style, fluency, and technical vocabulary usage—the scoring output from the LLM-as-judge became less stable and exhibited weaker correlation with expert judgment. Among these, the ``language style’’ dimension exhibited particularly high variance, as models often struggled to distinguish between stylistic appropriateness and technical correctness. Given the critical importance of factual accuracy in planning contexts, we ultimately excluded style-related factors from the final rubric to preserve evaluation precision. The incorporation of stylistic metrics is left for future refinement.

We also observed that concise prompts, when combined with few-shot scoring examples or structured reference answers such as annotated critical points, can significantly enhance the judgment accuracy and analytical capacity of the evaluation model.

To summarize, the prompt refinement process revealed that:
\begin{itemize}
    \item Evaluation prompts with fewer, well-defined dimensions produce more stable and reliable scores;
    \item Models with stronger grounding and reasoning capabilities exhibit greater resilience to prompt variation;
    \item Structured reference answers with annotated critical points enable small, cost-efficient judges such as GPT-4o-mini to achieve evaluation fidelity comparable to that of substantially larger models, establishing a favorable cost--accuracy tradeoff for large-scale benchmark evaluation.
\end{itemize}
These findings informed the final prompt templates used in the benchmark, striking a balance among evaluation fidelity, interpretability, and computational efficiency.

\subsection{VLM Performance in Planbench}
\label{sec:results}

\begin{figure}[htbp]
\centering
\includegraphics[width=0.8\linewidth]{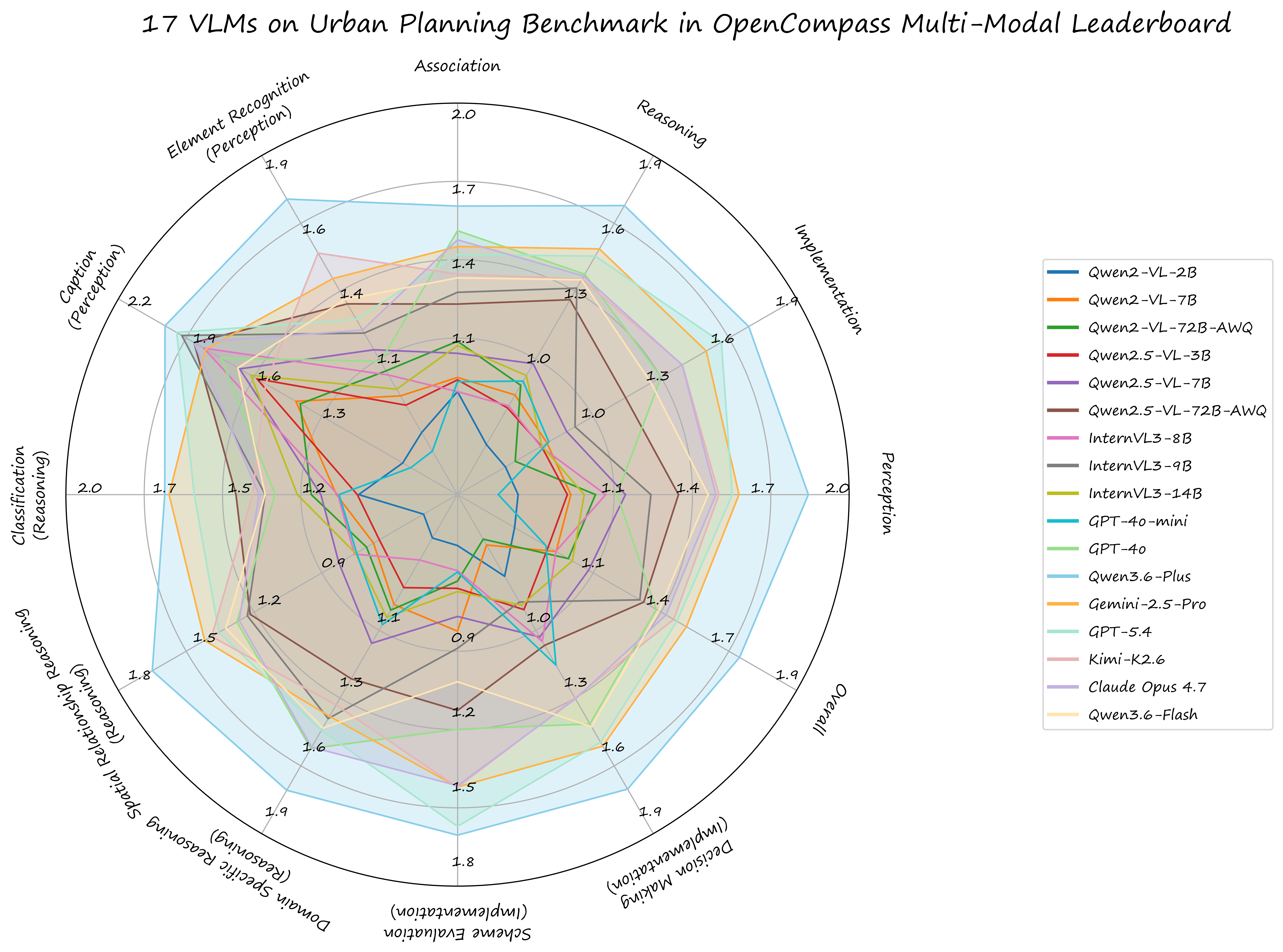}
\caption{Performance of 17 Models on Planbench-V Vision-Language Tasks across Two Generations}
\label{fig:result}
\end{figure}

To characterize both the current capability of VLMs and the trajectory of progress in planning map interpretation, we conducted two rounds of evaluation on the same 300-item stratified subset. The first round (2025) assessed 11 models representing the state of the art in vision-language pretraining at that time. The second round (2026) re-evaluated the identical items with six models that incorporate enhanced agentic reasoning capabilities. All models were scored using the same LLM-as-judge protocol described in Section~\ref{sec:prompt_refining}. Results for both rounds are presented jointly in Table~\ref{tab:model_performance} and Figure~\ref{fig:result}, with a separator line distinguishing the two generations.

\definecolor{best}{HTML}{E0F7FA}  
\definecolor{worst}{HTML}{FFEBEE} 

\newcommand{\best}[1]{\cellcolor{best}\textbf{#1}}
\newcommand{\worst}[1]{\cellcolor{worst}{#1}}

\begin{table*}[ht]
\centering
\caption{Performance of Models on PlanBench-V Vision-Language Tasks. Overall scores are computed on a 300-item stratified subset under the LLM-as-judge protocol described in Section~\ref{sec:prompt_refining}, and therefore are not the unweighted mean of the eight sub-task columns.}
\label{tab:model_performance}
\fontsize{6}{7}\selectfont
\setlength{\tabcolsep}{2pt}
\begin{tabular}{p{2.5cm}cccccccccc}
\toprule
\multirow{2}{*}{\textbf{Model}} & \multicolumn{2}{c}{\textbf{Perception}} & \multicolumn{3}{c}{\textbf{Reasoning}} & \multirow{2}{*}{\textbf{Association}} & \multicolumn{2}{c}{\textbf{Implementation}} & \multirow{2}{*}{\textbf{Overall}} & \multirow{2}{*}{\textbf{Rank}} \\
\cmidrule(r){2-3} \cmidrule(lr){4-6} \cmidrule(l){8-9}
& \shortstack{Elem.\\Recog.} & Caption & \shortstack{Classifi-\\cation} & \shortstack{Spatial\\Reasoning} & \shortstack{Domain\\Reasoning} & & \shortstack{Scheme\\Eval.} & \shortstack{Decision\\ making} & & \\
\midrule
\multicolumn{11}{c}{\textbf{First Round (2025)}} \\
\midrule
GPT-4o           & 1.051 & \best{1.878} & \best{1.406} & \best{1.305} & \best{1.564} & \best{1.527} & \best{1.223} & \best{1.429} & \best{1.342} & 1 \\
Qwen2.5-VL-72B\textsuperscript{a} & \best{1.299} & 1.825 & \best{1.406} & 1.248 & 1.263 & 1.253 & 1.153 & 1.090 & 1.288 & 2 \\
InternVL3-9B     & 1.173 & \best{1.878} & 1.297 & 1.260 & 1.435 & 1.297 & 0.921 & 0.903 & 1.271 & 3 \\
Qwen2.5-VL-7B    & 1.101 & 1.628 & 1.089 & 0.865 & 1.110 & 1.069 & 0.802 & 1.054 & 1.050 & 4 \\
InternVL3-14B    & 0.931 & 1.580 & 1.177 & 0.793 & 0.998 & 1.098 & 0.709 & 0.917 & 0.980 & 5 \\
Qwen2-VL-72B\textsuperscript{a} & 1.010 & 1.367 & 1.125 & 0.746 & 0.967 & 1.114 & 0.670 & \worst{0.632} & 0.963 & 6 \\
Qwen2-VL-7B      & 0.902 & 1.386 & 1.031 & 0.716 & 0.943 & 0.979 & 0.857 & 0.657 & 0.910 & 7 \\
InternVL3-8B     & 0.992 & 1.783 & 1.026 & 0.798 & 0.751 & \worst{0.926} & 0.631 & 1.073 & 0.909 & 8 \\
Qwen2.5-VL-3B    & 0.862 & 1.554 & 0.953 & 0.691 & 0.870 & 0.970 & 0.697 & 0.936 & 0.876 & 9 \\
GPT-4o-mini      & \worst{0.664} & \worst{0.890} & 1.021 & 0.789 & 1.030 & 0.963 & 0.636 & 1.175 & 0.866 & 10 \\
Qwen2-VL-2B      & 0.744 & 0.925 & \worst{0.948} & \worst{0.500} & \worst{0.656} & \worst{0.926} & \worst{0.537} & 0.792 & \worst{0.731} & 11 \\
\midrule
\multicolumn{11}{c}{\textbf{Second Round (2026)}} \\
\midrule
Qwen3.6-Plus     & \best{1.751} & \best{1.950} & \best{1.672} & \best{1.670} & \best{1.744} & \best{1.619} & \best{1.619} & \best{1.710} & \best{1.701} & 1 \\
Gemini-2.5-Pro   & 1.408 & 1.775 & 1.656 & 1.444 & 1.425 & 1.468 & 1.439 & 1.525 & 1.472 & 2 \\
GPT-5.4          & 1.233 & 1.900 & 1.562 & 1.383 & 1.486 & 1.438 & 1.586 & 1.508 & 1.431 & 3 \\
Kimi-K2.6        & 1.518 & \worst{1.516} & 1.344 & 1.414 & \worst{1.376} & 1.365 & 1.440 & \worst{1.317} & 1.417 & 4 \\
Claude-Opus-4.7  & \worst{1.186} & 1.825 & 1.320 & \worst{1.295} & 1.558 & 1.493 & 1.434 & 1.321 & 1.384 & 5 \\
Qwen3.6-Flash    & 1.318 & 1.636 & \worst{1.297} & 1.353 & 1.476 & \worst{1.351} & \worst{1.046} & 1.441 & \worst{1.353} & 6 \\
\bottomrule
\end{tabular}
\tabnote{
\begin{minipage}{1\linewidth}
\textsuperscript{a}Evaluated using the AWQ 4-bit quantized release (\texttt{Qwen2-VL-72B-Instruct-AWQ} and \texttt{Qwen2.5-VL-72B-Instruct-AWQ}); displayed without the \texttt{-AWQ} suffix for table readability.
\end{minipage}
}
\end{table*}

\subsubsection{First Round (2025)}

The first round evaluated 11 VLMs spanning a broad parameter range from 2B to 72B. The best-performing model was GPT-4o (Overall 1.342), followed by Qwen2.5-VL-72B-AWQ (1.288) and InternVL3-9B (1.271). Scaling offered diminishing returns: InternVL3-9B outperformed its larger counterpart InternVL3-14B (0.980), and flagship-scale models did not uniformly dominate smaller models. Detailed per-dimension analysis is provided below.

\subsubsection{Evaluation on Perception}
The \textbf{Perception} capability is assessed through two subtasks: \textbf{Element Recognition} and \textbf{Captioning}. Among all evaluated models in the first round, Qwen2.5-VL-72B-AWQ achieved the strongest Element Recognition score (1.299), while InternVL3-9B and GPT-4o tied for the highest Caption score (1.878). This result demonstrates robust visual grounding and scene interpretation with clear, domain-relevant terminology.

In contrast, GPT-4o-mini (0.664 in Element Recognition, 0.890 in Captioning) was the weakest performer on the two perception subtasks among the complete evaluations. The weakest small models struggled to correctly identify legend symbols and misattributed multiple design elements, particularly in crowded visual scenes. Their captions also exhibited vagueness and lacked domain-specific language. Notably, Qwen2-VL-2B showed limited overall capability with an average score of 0.731.

Overall, perception performance scales with model size up to a point, but several common failure patterns persist even in large models: color misclassification, hallucination of objects absent from the image, and degraded reliability on crowded or multi-element scenes.

\subsubsection{Evaluation on Reasoning}
The \textbf{Reasoning} capability consists of three subtasks: \textbf{Classification}, \textbf{Spatial Relationship Reasoning}, and \textbf{Domain-Specific Reasoning}. This dimension assesses the ability of a model to categorize urban elements accurately, understand their spatial relations, and reason with planning knowledge.

The top performer in this dimension was GPT-4o, achieving an overall Reasoning score of 1.425. It attained 1.406 in Classification, 1.305 in Spatial Reasoning, and 1.564 in Domain-Specific Reasoning---the highest score among first-round models in the latter category. These results demonstrate the robust capacity of GPT-4o to conduct structured inferences based on visual layouts and implicit planning constraints.

Following closely were InternVL3-9B (1.331) and Qwen2.5-VL-72B-AWQ (1.306). InternVL3-9B performed particularly strongly in Domain-Specific Reasoning (1.435) and Classification (1.297), while maintaining solid Spatial Reasoning (1.260). Qwen2.5-VL-72B-AWQ achieved the highest Classification score (1.406, tied with GPT-4o) and exhibited balanced performance across all three subtasks, although its Spatial Reasoning (1.248) lagged slightly behind that of InternVL3-9B.

In contrast, the weakest performer was Qwen2-VL-2B, with a Reasoning score of 0.701, followed by Qwen2.5-VL-3B (0.838). Qwen2.5-VL-3B struggled most notably in Spatial Reasoning (0.691), indicating fundamental challenges with interpreting spatial relationships in complex urban layouts. Qwen2-VL-7B also showed limitations with an overall score of 0.897, particularly in Spatial Reasoning (0.716) and Classification (1.031).

The reasoning tasks revealed that even high-scoring models encountered difficulty with multi-step logic and condition-based classification. Qwen2.5-VL-72B-Instruct-AWQ frequently produced overgeneralized answers or missed key spatial and policy conditions, leading to incorrect or superficial classifications.

\subsubsection{Evaluation on Association}
The \textbf{Association} dimension focuses on the ability of a model to link visual elements and spatial design with relevant policy knowledge, specifically evaluated through the Policy Association task. This dimension tests whether a model can identify and align visual evidence with appropriate regulatory frameworks and planning categories, a critical capability in professional design evaluation.

The best performer in this dimension was GPT-4o, achieving a score of 1.527. It was closely followed by InternVL3-9B (1.297) and Qwen2.5-VL-72B-AWQ (1.253). These models demonstrated the capacity to cite relevant planning documents, recognize administrative hierarchies such as national versus local regulations, and align visual cues with textual policy language. For instance, in urban renewal scenarios, these models correctly referenced floor-area-ratio regulations or ecological redline zones based on spatial overlays in the input images.

Despite this, even the strongest models exhibited limited policy awareness in terms of structural specificity. Several answers listed relevant policies but failed to clarify their jurisdictional scope or functional classification. A recurring issue was category confusion, in which local planning guidelines, special-purpose plans, and national strategic frameworks were intermingled without proper distinction. This lack of differentiation weakens the reliability of the association chain and limits its applicability in scenario-based reasoning.

The lowest-scoring models in this dimension were InternVL3-8B and Qwen2-VL-2B (both 0.926), followed by GPT-4o-mini (0.963) and Qwen2.5-VL-3B (0.970). These models frequently hallucinated plausible-sounding regulations with no basis in the image or failed to distinguish between visually similar zones, for instance confusing green buffer zones with agricultural protection areas. In addition, they often emphasized advantages such as innovation or aesthetics but neglected to identify regulatory risks or constraints, reflecting a one-sided interpretation of planning implications.

Association tasks required models to connect visual evidence with planning policies. Despite high surface-level fluency, models lacked \textbf{policy awareness} and \textbf{structural coherence}: Qwen2.5-VL-72B, for instance, often listed multiple policy items but failed to distinguish \textbf{administrative levels} or \textbf{planning categories}, resulting in conceptual overlaps and ambiguity. Models also tended to highlight strengths such as innovation or leadership while neglecting limitations, and produced syntactically coherent but unsupported reasoning chains.

\subsubsection{Evaluation on Implementation}
The \textbf{Implementation} dimension assesses how well models can generate actionable feedback and evaluation statements within professional design contexts. It comprises two tasks: \textbf{Scheme Evaluation} and \textbf{Decision-Making}, each requiring models to analyze spatial proposals and produce concise, domain-appropriate responses aligned with planning logic.

The best performer was GPT-4o, achieving 1.223 in Scheme Evaluation and 1.429 in Decision-Making. Its answers were well-structured, referenced key urban design metrics such as density and accessibility, and reflected trade-offs among aesthetics, ecology, and policy feasibility.

Following closely were Qwen2.5-VL-72B-AWQ (1.153 and 1.090) and InternVL3-9B (0.921 and 0.903). Qwen2.5-VL-72B-AWQ demonstrated balanced performance across both tasks, while InternVL3-9B exhibited particular strength in Scheme Evaluation but relatively weaker capability in Decision-Making.

Nevertheless, common issues persisted across models. Qwen2.5-VL-72B-Instruct, despite its large parameter count, produced verbose and unfocused responses that often lacked prioritization or avoided value judgments altogether. Many models failed to identify constraints such as zoning incompatibilities or ecological conflicts---central to real-world planning evaluations---and Qwen2-VL-2B recorded the lowest overall average (0.731), consistently failing to deliver actionable, terminology-grounded advice.

\subsubsection{Second Round (2026)}

The second round evaluated six models that incorporate enhanced agentic reasoning capabilities. Results are shown in the lower panel of Table~\ref{tab:model_performance}. The standout performer is Qwen3.6-Plus, which achieves an Overall score of 1.701, decisively surpassing all other models in both generations. Its dominance is broad: it ranks first in every task category, with particularly commanding leads in Element Recognition (1.751 vs.\ the next-best Kimi-K2.6 at 1.518) and Scheme Evaluation (1.619 vs.\ the next-best GPT-5.4 at 1.586).

The remaining 2026 models cluster within a narrower band (1.353--1.472). Gemini-2.5-Pro (1.472) and GPT-5.4 (1.431) form the second tier. Gemini-2.5-Pro excels in Classification (1.656), while GPT-5.4 achieves the second-highest Caption score (1.900, after Qwen3.6-Plus at 1.950) and the second-highest Scheme Evaluation score (1.586), but is held back by comparatively weak Element Recognition (1.233). Kimi-K2.6 (1.417) is notable for strong Element Recognition (1.518). Claude-Opus-4.7 (1.384) presents a polarized profile: strong on Domain-Specific Reasoning (1.558) and Policy Association (1.493), yet the weakest in Element Recognition (1.186). At the lower end, Qwen3.6-Flash (1.353) performs creditably for a lightweight model, but its Scheme Evaluation score (1.046) is the weakest among all 2026 models.

\subsubsection{Cross-Generation Comparison}

Comparing the two rounds reveals a clear step-change in capability. The best 2026 model (Qwen3.6-Plus, 1.701) outperforms the best 2025 model (GPT-4o, 1.342) by a margin of 0.359, an improvement of approximately 27\%. Strikingly, even the lightest 2026 model, Qwen3.6-Flash (1.353), exceeds the top 2025 score. This pattern confirms that the agentic reasoning enhancements present in the 2026 generation yield a substantive and uniform upward shift across the entire capability spectrum.

The gains are not uniform across dimensions. The most dramatic improvement occurs in Scheme Evaluation, where the 2026 generation mean (1.427) is 77.7\% higher than that of the 2025 generation (0.803), reflecting the capacity of agentic models to structure multi-criteria judgments and produce actionable planning critiques. Spatial Reasoning (0.883 $\rightarrow$ 1.426, +0.543) and Decision-making (0.969 $\rightarrow$ 1.470, +0.501) also show large absolute gains. Domain-Specific Reasoning (1.053 $\rightarrow$ 1.511, +0.458) and Element Recognition (0.975 $\rightarrow$ 1.402, +0.427) improve substantially, while Policy Association (1.102 $\rightarrow$ 1.456, +0.354), Classification (1.134 $\rightarrow$ 1.475, +0.341), and Captioning (1.518 $\rightarrow$ 1.767, +0.249) show smaller absolute gains.

\subsection{Image Resolution Sensitivity Analysis}
\label{sec:resolution}

A natural concern when evaluating VLMs on planning maps is whether variation in image resolution systematically biases model scores, given that the benchmark draws images from heterogeneous sources whose total pixel counts vary by roughly 46-fold. To investigate this, we conducted a resolution sensitivity analysis on the evaluation subset, which contains 63 unique images with resolutions ranging from 352$\times$373 to 2,057$\times$2,953 pixels.

\begin{figure}[htbp]
\centering
\includegraphics[width=0.8\linewidth]{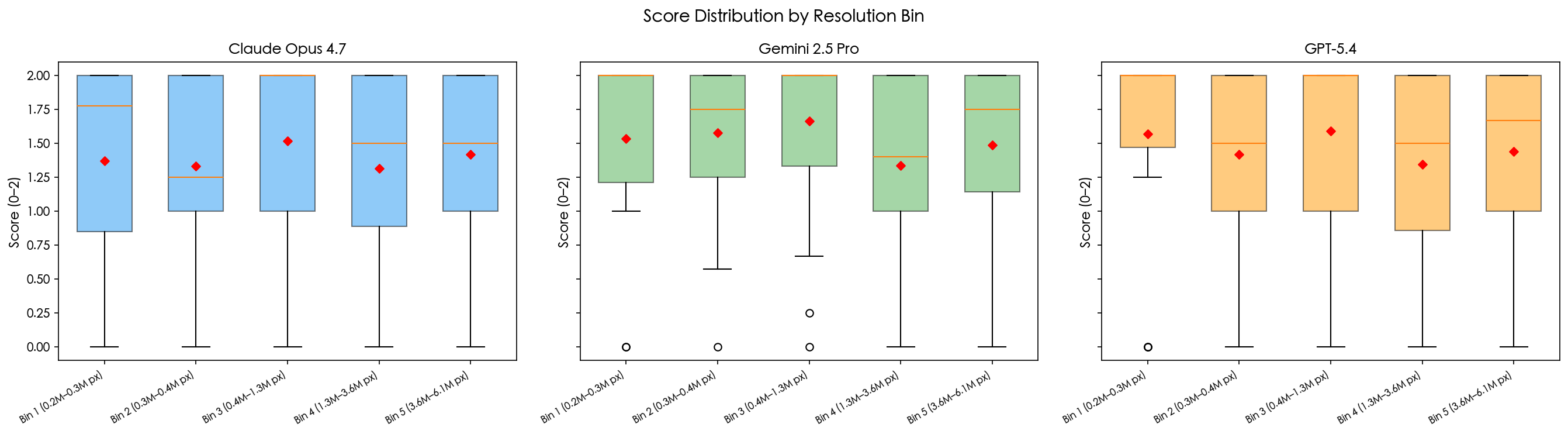}
\caption{Image resolution sensitivity analysis. Boxplots of per-model scores aggregated into five quantile resolution bins. No meaningful or statistically significant difference is observed across bins for any model (ANOVA, all $p > 0.05$).}
\label{fig:resolution}
\end{figure}

For this analysis, we scored the subset across three representative models (Claude Opus 4.7, Gemini 2.5 Pro, and GPT-5.4) and computed Spearman rank correlations between per-item scores and six resolution metrics: log\textsubscript{10}(total pixels), total pixels, $\sqrt{\text{pixels}}$, width, height, and aspect ratio. As shown in Figure~\ref{fig:resolution}, none of the models exhibited a meaningful correlation between image resolution and score. The strongest per-model correlation was observed for Claude Opus 4.7 on width ($\rho = +0.057$, $p = 0.329$), and all log\textsubscript{10}(pixels) correlations remained negligible ($|\rho| < 0.06$, $p > 0.30$ across all three models). We further partitioned the images into five quantile bins by log\textsubscript{10}(pixels) and conducted one-way ANOVA tests for each model; no model showed significant between-bin score differences (all $p > 0.05$). Aggregating scores to the image level yielded a Spearman $\rho$ of $-0.222$ ($p = 0.080$).

These results collectively indicate that image resolution does not act as a confounding factor in the evaluation of PlanBench-V. The natural variation in map image dimensions within the benchmark is sufficiently narrow and uncorrelated with task difficulty that it does not introduce systematic bias into model rankings. This finding supports the validity of cross-model comparisons reported in Section~\ref{sec:results} without requiring resolution normalization.

\section{Conclusion}
This study presents PlanBench-V, the first comprehensive benchmark for evaluating the capabilities of VLMs in spatial planning map interpretation. We constructed the SPMD with 223 planning maps and 1,629 expert-annotated QA pairs, organized across four progressive dimensions—Perception, Reasoning, Association, and Implementation—that reflect the full cognitive pipeline of planning practice. Through two rounds of evaluation spanning two generations of models, we characterized both the current state of VLM capability and the trajectory of progress. The best 2025 model, GPT-4o, achieved an Overall score of 1.342; the best 2026 agentic reasoning model, Qwen3.6-Plus, improved this to 1.701, a margin of 0.359 or approximately 27\%. Strikingly, even the lightest 2026 model exceeds the best 2025 score, confirming that agentic reasoning yields a substantive generation-level uplift.

However, despite this progress, implementation-oriented tasks remain a central challenge. Even the best 2026 models can still struggle to generate concise, actionable, and well-structured evaluations, often producing verbose outputs with limited domain specificity. The absolute gains are markedly uneven: Scheme Evaluation, Spatial Reasoning, Decision-making, Domain-Specific Reasoning, and Element Recognition show the largest leaps, while Captioning, Classification, and Policy Association improve less. This pattern suggests that general-purpose reasoning enhancements alone cannot compensate for the lack of structured domain knowledge in planning-specific tasks.

A plausible explanation is that implementation-oriented tasks impose a substantially higher cognitive burden than conventional VQA problems. They demand long-horizon reasoning within a single query, requiring the simultaneous integration of visual perception, spatial relationship understanding, domain-specific planning knowledge, and policy-context constraints. The resulting challenge arises not from the number of questions but from the depth of reasoning and the heterogeneity of information that must be coordinated within a continuous inference process. Even the strongest agentic models exhibit characteristic failure patterns, including unfocused verbosity, drifting attention, insufficient policy grounding, and inconsistencies across reasoning stages.

These observations confirm that the performance bottleneck in planning map interpretation is not merely a consequence of missing domain knowledge. Rather, it reflects a fundamental mismatch between prevailing vision-language learning paradigms and the cognitive demands of professional planning practice, which requires normative judgment under legal frameworks, hierarchical governance structures, and competing public objectives.

Future research should therefore pursue domain-adaptive multimodal learning strategies that explicitly encode planning logic, regulatory constraints, and spatial decision-making processes. Integrating structured policy knowledge, constraint-aware reasoning mechanisms, and theory-informed supervision may prove essential for bridging the gap between general-purpose multimodal intelligence and expert-level urban planning cognition. By articulating these challenges across two model generations, PlanBench-V provides not only an evaluation tool but also a conceptual foundation for advancing vision-language models toward more reliable and responsible applications in spatial governance.

\bibliography{interacttfvsample}

@book{palmer1999vision,
  title     = {Vision Science: Photons to Phenomenology},
  author    = {Palmer, Stephen E.},
  year      = {1999},
  publisher = {MIT Press}
}

@book{koffka1935principles,
  title     = {Principles of Gestalt Psychology},
  author    = {Koffka, Kurt},
  year      = {1935},
  publisher = {Harcourt, Brace and Company}
}

@article{wertheimer1938gestalt,
  title   = {Gestalt Theory},
  author  = {Wertheimer, Max},
  journal = {Social Research},
  volume  = {5},
  number  = {1},
  pages   = {81--121},
  year    = {1938}
}

@book{mayer2005cambridge,
  title     = {The Cambridge Handbook of Multimedia Learning},
  author    = {Mayer, Richard E.},
  year      = {2005},
  publisher = {Cambridge University Press}
}

@book{arnheim1969visual,
  title     = {Visual Thinking},
  author    = {Arnheim, Rudolf},
  year      = {1969},
  publisher = {University of California Press}
}

@book{alexander1977pattern,
  title     = {A Pattern Language: Towns, Buildings, Construction},
  author    = {Alexander, Christopher},
  year      = {1977},
  publisher = {Oxford University Press}
}

@book{kress2006reading,
  title     = {Reading Images: The Grammar of Visual Design},
  author    = {Kress, Gunther and van Leeuwen, Theo},
  edition   = {2nd},
  year      = {2006},
  publisher = {Routledge}
}

@book{bertin1983semiology,
  title     = {Semiology of Graphics: Diagrams, Networks, Maps},
  author    = {Bertin, Jacques},
  year      = {1983},
  publisher = {University of Wisconsin Press},
  translator = {Berg, W. J.}
}

@article{moroni2017graphic,
  title   = {Graphic Rules in Planning: A Critical Exploration of Normative Drawings Starting from Zoning Maps and Form-Based Codes},
  author  = {Moroni, Stefano and Lorini, Giandomenico},
  journal = {Planning Theory},
  volume  = {16},
  number  = {3},
  pages   = {318--338},
  year    = {2017},
  doi     = {10.1177/1473095216672342}
}

@article{li2025communicative,
  title     = {The communicative turn in planning? Examining community planner's role as a third actor in Beijing, China},
  author    = {Li, Zhen and Lin, Yanliu and Hooimeijer, Pieter and Monstadt, Jochen and He, Junyao},
  journal   = {Cities},
  volume    = {159},
  year      = {2025},
  pages     = {105785},
  issn      = {0264-2751},
  doi       = {10.1016/j.cities.2025.105785},
  url       = {https://www.sciencedirect.com/science/article/pii/S026427512500085X}
}

@book{duhr2007visual,
  title     = {The Visual Language of Spatial Planning: Exploring Cartographic Representations for Spatial Planning in Europe},
  author    = {D{\"u}hr, Stefanie},
  year      = {2007},
  edition   = {1st},
  publisher = {Routledge},
  address   = {London},
  doi       = {10.4324/9780203965818},
  isbn      = {9780203965818},
  url       = {https://doi.org/10.4324/9780203965818}
}

@incollection{duhr2009visualising,
  author    = {Dühr, Stefanie},
  title     = {Visualising Spatial Policy in Europe},
  booktitle = {Planning Cultures in Europe},
  editor    = {Othengrafen, Frank and Knieling, Joerg},
  year      = {2009},
  publisher = {Routledge},
  pages     = {24},
  edition   = {1st},
  address   = {London},
  isbn      = {9781315246727},
  doi       = {10.4324/9781315246727}
}

@incollection{burgess2009shift,
  title={The shift from master planning to strategic planning},
  author={Burgess, Rod and Carmona, Magdalena I.},
  booktitle={Planning through projects: Moving from master planning to strategic planning - 30 cities},
  editor={Carmona, Magdalena I. and Burgess, Rod and Badenhorst, M. S.},
  pages={12--42},
  year={2009},
  publisher={Technepress}
}

@article{faludi2000esdp,
  title={The European Spatial Development Perspective--What Next?},
  author={Faludi, Andreas},
  journal={European Planning Studies},
  volume={8},
  number={2},
  pages={237--250},
  year={2000},
  publisher={Taylor \& Francis},
  doi={10.1080/713666411}
}

@article{albrechts2003strategic,
  title={Strategic Spatial Planning and Regional Governance in Europe},
  author={Albrechts, Louis and Healey, Patsy and Kunzmann, Klaus R.},
  journal={Journal of the American Planning Association},
  volume={69},
  number={2},
  pages={113--129},
  year={2003},
  publisher={Taylor \& Francis},
  doi={10.1080/01944360308976301},
  issn={0194-4363}
}

@article{harley1989deconstructing,
  title={Deconstructing the Map},
  author={Harley, J. B.},
  journal={Cartographica: The International Journal for Geographic Information and Geovisualization},
  volume={26},
  number={2},
  pages={1--20},
  year={1989},
  publisher={University of Toronto Press},
  doi={10.3138/J674-7727-1757-1412}
}

@book{harley2002new,
  title={The New Nature of Maps: Essays in the History of Cartography},
  author={Harley, J. B.},
  editor={Laxton, Paul},
  year={2002},
  publisher={Johns Hopkins University Press},
  address={Baltimore}
}

@book{lynch1984,
  title={Good city form},
  author={Lynch, Kevin},
  year={1984},
  publisher={MIT press}
}

@article{steinitz1995,
  title={A framework for theory and practice in landscape planning},
  author={Steinitz, Carl},
  journal={Process architecture},
  volume={127},
  pages={12--31},
  year={1995}
}

@book{healey1997,
  title={Collaborative planning: Shaping places in fragmented societies},
  author={Healey, Patsy},
  year={1997},
  publisher={UBC Press}
}

@misc{bai2023qwen,
      title={Qwen-VL: A Versatile Vision-Language Model for Understanding, Localization, Text Reading, and Beyond}, 
      author={Jinze Bai and Shuai Bai and Shusheng Yang and Shijie Wang and Sinan Tan and Peng Wang and Junyang Lin and Chang Zhou and Jingren Zhou},
      year={2023},
      eprint={2308.12966},
      archivePrefix={arXiv},
      primaryClass={cs.CV}
}

@article{chen2023internvl,
  title={InternVL: Scaling up Vision Foundation Models and Aligning for Generic Visual-Linguistic Tasks},
  author={Chen, Zhe and Wu, Jiannan and Wang, Wenhai and Su, Weijie and Chen, Guo and Xing, Sen and Zhong, Muyan and Zhang, Qinglong and Zhu, Xizhou and Lu, Lewei and Li, Bin and Luo, Ping and Lu, Tong and Qiao, Yu and Dai, Jifeng},
  journal={arXiv preprint arXiv:2312.14238},
  year={2023}
}

@inproceedings{yue2023mmmu,
      title={{MMMU}: A Massive Multi-discipline Multimodal Understanding and Reasoning Benchmark for Expert AGI}, 
      author={Xiang Yue and Yuval Ben-David and Sivan Nova and Xingjian Shi and Ka-Wai Kid Lin and Badr Al-Ghamdi and Pujung Kim and Ryan Roggener and Dror Glickman and Ser-Nam Lim and Reid Pryzant and Wen-tau Yih and Yair Carmon and Wentao Wang and C.J. Richard Shi},
      booktitle={Proceedings of the IEEE/CVF Conference on Computer Vision and Pattern Recognition (CVPR)},
      year={2024},
      eprint={2311.16502},
      archivePrefix={arXiv},
      primaryClass={cs.CV}
}

@inproceedings{zhu-etal-2025-plangpt,
    title = "{P}lan{GPT}: Enhancing Urban Planning with a Tailored Agent Framework",
    author = "Zhu, He and Chen, Guanhua and Zhang, Wenjia",
    editor = "Rehm, Georg and Li, Yunyao",
    booktitle = "Proceedings of the 63rd Annual Meeting of the Association for Computational Linguistics (Volume 6: Industry Track)",
    month = jul,
    year = "2025",
    address = "Vienna, Austria",
    publisher = "Association for Computational Linguistics",
    url = "https://aclanthology.org/2025.acl-industry.54/",
    doi = "10.18653/v1/2025.acl-industry.54",
    pages = "764--783",
    ISBN = "979-8-89176-288-6",
}

@misc{zhu2025plangptvlenhancingurbanplanning,
      title={PlanGPT-VL: Enhancing Urban Planning with Domain-Specific Vision-Language Models}, 
      author={He Zhu and Junyou Su and Minxin Chen and Wen Wang and Yijie Deng and Guanhua Chen and Wenjia Zhang},
      year={2025},
      eprint={2505.14481},
      archivePrefix={arXiv},
      primaryClass={cs.CL},
      url={https://arxiv.org/abs/2505.14481}, 
}

@misc{chang2022mapqadatasetquestionanswering,
      title={MapQA: A Dataset for Question Answering on Choropleth Maps}, 
      author={Shuaichen Chang and David Palzer and Jialin Li and Eric Fosler-Lussier and Ningchuan Xiao},
      year={2022},
      eprint={2211.08545},
      archivePrefix={arXiv},
      primaryClass={cs.CV},
      url={https://arxiv.org/abs/2211.08545}, 
}

@article{roberts2023charting,
  title={{Charting New Territories: Exploring the geographic and geospatial capabilities of multimodal LLMs}},
  author={Roberts, Jonathan and L{\"u}ddecke, Timo and Sheikh, Rehan and Han, Kai and Albanie, Samuel},
  journal={arXiv preprint arXiv:2311.14656},
  year={2023}
}

@misc{agrawal_vqa_2016,
	title = {{VQA}: {Visual} {Question} {Answering}},
	shorttitle = {{VQA}},
	url = {http://arxiv.org/abs/1505.00468},
	doi = {10.48550/arXiv.1505.00468},
	urldate = {2025-07-23},
	publisher = {arXiv},
	author = {Agrawal, Aishwarya and Lu, Jiasen and Antol, Stanislaw and Mitchell, Margaret and Zitnick, C. Lawrence and Batra, Dhruv and Parikh, Devi},
	month = oct,
	year = {2016},
	note = {arXiv:1505.00468 [cs]},
	}

@misc{lin_microsoft_2015,
	title = {Microsoft {COCO}: {Common} {Objects} in {Context}},
	shorttitle = {Microsoft {COCO}},
	url = {http://arxiv.org/abs/1405.0312},
	doi = {10.48550/arXiv.1405.0312},
	urldate = {2025-07-23},
	publisher = {arXiv},
	author = {Lin, Tsung-Yi and Maire, Michael and Belongie, Serge and Bourdev, Lubomir and Girshick, Ross and Hays, James and Perona, Pietro and Ramanan, Deva and Zitnick, C. Lawrence and Dollár, Piotr},
	month = feb,
	year = {2015},
	note = {arXiv:1405.0312 [cs]},
	}

@article{krishna_visual_2017,
	title = {Visual {Genome}: {Connecting} {Language} and {Vision} {Using} {Crowdsourced} {Dense} {Image} {Annotations}},
	volume = {123},
	issn = {1573-1405},
	shorttitle = {Visual {Genome}},
	url = {https://doi.org/10.1007/s11263-016-0981-7},
	doi = {10.1007/s11263-016-0981-7},
	number = {1},
	urldate = {2025-07-23},
	journal = {International Journal of Computer Vision},
	author = {Krishna, Ranjay and Zhu, Yuke and Groth, Oliver and Johnson, Justin and Hata, Kenji and Kravitz, Joshua and Chen, Stephanie and Kalantidis, Yannis and Li, Li-Jia and Shamma, David A. and Bernstein, Michael S. and Fei-Fei, Li},
	month = may,
	year = {2017},
	pages = {32--73},
	}

@misc{wang_needle_2024,
	title = {Needle {In} {A} {Multimodal} {Haystack}},
	url = {http://arxiv.org/abs/2406.07230},
	doi = {10.48550/arXiv.2406.07230},
	urldate = {2025-07-23},
	publisher = {arXiv},
	author = {Wang, Weiyun and Zhang, Shuibo and Ren, Yiming and Duan, Yuchen and Li, Tiantong and Liu, Shuo and Hu, Mengkang and Chen, Zhe and Zhang, Kaipeng and Lu, Lewei and Zhu, Xizhou and Luo, Ping and Qiao, Yu and Dai, Jifeng and Shao, Wenqi and Wang, Wenhai},
	month = oct,
	year = {2024},
	note = {arXiv:2406.07230 [cs]},
	}

@misc{mathew2021documentvisualquestionanswering,
      title={Document Visual Question Answering Challenge 2020}, 
      author={Minesh Mathew and Ruben Tito and Dimosthenis Karatzas and R. Manmatha and C. V. Jawahar},
      year={2021},
      eprint={2008.08899},
      archivePrefix={arXiv},
      primaryClass={cs.CV},
      url={https://arxiv.org/abs/2008.08899}, 
}

@misc{li2023scigraphqalargescalesyntheticmultiturn,
      title={SciGraphQA: A Large-Scale Synthetic Multi-Turn Question-Answering Dataset for Scientific Graphs}, 
      author={Shengzhi Li and Nima Tajbakhsh},
      year={2023},
      eprint={2308.03349},
      archivePrefix={arXiv},
      primaryClass={cs.CL},
      url={https://arxiv.org/abs/2308.03349}, 
}

@misc{roberts2024scifibenchbenchmarkinglargemultimodal,
      title={SciFIBench: Benchmarking Large Multimodal Models for Scientific Figure Interpretation}, 
      author={Jonathan Roberts and Kai Han and Neil Houlsby and Samuel Albanie},
      year={2024},
      eprint={2405.08807},
      archivePrefix={arXiv},
      primaryClass={cs.CV},
      url={https://arxiv.org/abs/2405.08807}, 
}

@misc{doris2024DesignQAMultimodal,
  title = {{{DesignQA}}: {{A Multimodal Benchmark}} for {{Evaluating Large Language Models}}' {{Understanding}} of {{Engineering Documentation}}},
  author = {Doris, Anna C. and Grandi, Daniele and Tomich, Ryan and Alam, Md Ferdous and Ataei, Mohammadmehdi and Cheong, Hyunmin and Ahmed, Faez},
  year = {2024},
  month = aug,
  number = {arXiv:2404.07917},
  eprint = {2404.07917},
  primaryclass = {cs},
  publisher = {arXiv},
  doi = {10.48550/arXiv.2404.07917},
  urldate = {2025-05-13},
  archiveprefix = {arXiv}
}

@article{rautenbach2017Developmentevaluation,
  title = {Development and Evaluation of a Specialized Task Taxonomy for Spatial Planning -- {{A}} Map Literacy Experiment with Topographic Maps},
  author = {Rautenbach, Victoria and Coetzee, Serena and {\c C}{\"o}ltekin, Arzu},
  year = {2017},
  month = may,
  journal = {ISPRS Journal of Photogrammetry and Remote Sensing},
  volume = {127},
  pages = {16--26},
  publisher = {Elsevier},
  doi = {10.1016/j.isprsjprs.2016.06.013},
  urldate = {2025-05-13},
  langid = {american}
}

@article{li2023llava,
  title={Llava-med: Training a large language-and-vision assistant for biomedicine in one day},
  author={Li, Chunyuan and Wong, Cliff and Zhang, Sheng and Usuyama, Naoto and Liu, Haotian and Yang, Jianwei and Naumann, Tristan and Poon, Hoifung and Gao, Jianfeng},
  journal={Advances in Neural Information Processing Systems},
  volume={36},
  pages={28541--28564},
  year={2023}
}

@article{lai2025med,
  title={Med-r1: Reinforcement learning for generalizable medical reasoning in vision-language models},
  author={Lai, Yuxiang and Zhong, Jike and Li, Ming and Zhao, Shitian and Yang, Xiaofeng},
  journal={arXiv preprint arXiv:2503.13939},
  year={2025}
}

@article{pan2025medvlm,
  title={Medvlm-r1: Incentivizing medical reasoning capability of vision-language models (vlms) via reinforcement learning},
  author={Pan, Jiazhen and Liu, Che and Wu, Junde and Liu, Fenglin and Zhu, Jiayuan and Li, Hongwei Bran and Chen, Chen and Ouyang, Cheng and Rueckert, Daniel},
  journal={arXiv preprint arXiv:2502.19634},
  year={2025}
}

@article{zhang2024bb,
  title={BB-GeoGPT: A framework for learning a large language model for geographic information science},
  author={Zhang, Yifan and Wang, Zhiyun and He, Zhengting and Li, Jingxuan and Mai, Gengchen and Lin, Jianfeng and Wei, Cheng and Yu, Wenhao},
  journal={Information Processing \& Management},
  volume={61},
  number={5},
  pages={103808},
  year={2024},
  publisher={Elsevier}
}

@article{zhang2024mapgpt,
  title={MapGPT: an autonomous framework for mapping by integrating large language model and cartographic tools},
  author={Zhang, Yifan and He, Zhengting and Li, Jingxuan and Lin, Jianfeng and Guan, Qingfeng and Yu, Wenhao},
  journal={Cartography and Geographic Information Science},
  volume={51},
  number={6},
  pages={717--743},
  year={2024},
  publisher={Taylor \& Francis}
}

@article{chen2025bring,
  title={Bring Reason to Vision: Understanding Perception and Reasoning through Model Merging},
  author={Chen, Shiqi and Zhang, Jinghan and Zhu, Tongyao and Liu, Wei and Gao, Siyang and Xiong, Miao and Li, Manling and He, Junxian},
  journal={arXiv preprint arXiv:2505.05464},
  year={2025}
}

@article{shen2025vlm,
  title={Vlm-r1: A stable and generalizable r1-style large vision-language model},
  author={Shen, Haozhan and Liu, Peng and Li, Jingcheng and Fang, Chunxin and Ma, Yibo and Liao, Jiajia and Shen, Qiaoli and Zhang, Zilun and Zhao, Kangjia and Zhang, Qianqian and others},
  journal={arXiv preprint arXiv:2504.07615},
  year={2025}
}

@article{zhu2025internvl3,
  title={InternVL3: Exploring Advanced Training and Test-Time Recipes for Open-Source Multimodal Models},
  author={Zhu, Jinguo and Wang, Weiyun and Chen, Zhe and Liu, Zhaoyang and Ye, Shenglong and Gu, Lixin and Duan, Yuchen and Tian, Hao and Su, Weijie and Shao, Jie and others},
  journal={arXiv preprint arXiv:2504.10479},
  year={2025}
}

@article{huang2024aesbench,
  title={Aesbench: An expert benchmark for multimodal large language models on image aesthetics perception},
  author={Huang, Yipo and Yuan, Quan and Sheng, Xiangfei and Yang, Zhichao and Wu, Haoning and Chen, Pengfei and Yang, Yuzhe and Li, Leida and Lin, Weisi},
  journal={arXiv preprint arXiv:2401.08276},
  year={2024}
}

@article{lin2024designprobe,
  title={Designprobe: A graphic design benchmark for multimodal large language models},
  author={Lin, Jieru and Huang, Danqing and Zhao, Tiejun and Zhan, Dechen and Lin, Chin-Yew},
  journal={arXiv preprint arXiv:2404.14801},
  year={2024}
}

@article{zhou2024uniaa,
  title={Uniaa: A unified multi-modal image aesthetic assessment baseline and benchmark},
  author={Zhou, Zhaokun and Wang, Qiulin and Lin, Bin and Su, Yiwei and Chen, Rui and Tao, Xin and Zheng, Amin and Yuan, Li and Wan, Pengfei and Zhang, Di},
  journal={arXiv preprint arXiv:2404.09619},
  year={2024}
}

@inproceedings{qian2024nuscenes,
  title={Nuscenes-qa: A multi-modal visual question answering benchmark for autonomous driving scenario},
  author={Qian, Tianwen and Chen, Jingjing and Zhuo, Linhai and Jiao, Yang and Jiang, Yu-Gang},
  booktitle={Proceedings of the AAAI Conference on Artificial Intelligence},
  volume={38},
  number={5},
  pages={4542--4550},
  year={2024}
}

@inproceedings{sima2024drivelm,
  title={Drivelm: Driving with graph visual question answering},
  author={Sima, Chonghao and Renz, Katrin and Chitta, Kashyap and Chen, Li and Zhang, Hanxue and Xie, Chengen and Bei{\ss}wenger, Jens and Luo, Ping and Geiger, Andreas and Li, Hongyang},
  booktitle={European Conference on Computer Vision},
  pages={256--274},
  year={2024},
  organization={Springer}
}

@misc{wang2024qwen2vl,
      title={Qwen2-VL: Enhancing Vision-Language Model's Perception of the World at Any Resolution}, 
      author={Peng Wang and Shuai Bai and Sinan Tan and Shijie Wang and Zhihao Fan and Jinze Bai and Keqin Chen and Xuejing Liu and Jialin Wang and Wenbin Ge and Yang Fan and Kai Dang and Mengfei Du and Xuancheng Ren and Rui Men and Dayiheng Liu and Chang Zhou and Jingren Zhou and Junyang Lin},
      year={2024},
      eprint={2409.12191},
      archivePrefix={arXiv},
      primaryClass={cs.CV},
      url={https://arxiv.org/abs/2409.12191}, 
}

@misc{liu2023llava,
      title={Visual Instruction Tuning}, 
      author={Haotian Liu and Chunyuan Li and Qingyang Wu and Yong Jae Lee},
      year={2023},
      eprint={2304.08485},
      archivePrefix={arXiv},
      primaryClass={cs.CV},
      url={https://arxiv.org/abs/2304.08485}, 
}

@article{guo2025seed1,
  title={Seed1. 5-VL Technical Report},
  author={Guo, Dong and Wu, Faming and Zhu, Feida and Leng, Fuxing and Shi, Guang and Chen, Haobin and Fan, Haoqi and Wang, Jian and Jiang, Jianyu and Wang, Jiawei and others},
  journal={arXiv preprint arXiv:2505.07062},
  year={2025}
}

@article{gpt4o,
  title={Gpt-4o system card},
  author={Hurst, Aaron and Lerer, Adam and Goucher, Adam P and Perelman, Adam and Ramesh, Aditya and Clark, Aidan and Ostrow, AJ and Welihinda, Akila and Hayes, Alan and Radford, Alec and others},
  journal={arXiv preprint arXiv:2410.21276},
  year={2024}
}

@misc{gemini,
  author = {Google DeepMind},
  title = {Gemini},
  year = {2023},
  howpublished = {\url{https://gemini.google.com}},
}

\end{document}